\documentclass[11pt,a4paper]{article}
\usepackage[hyperref]{acl2017}
\usepackage{times}
\usepackage{latexsym}
\usepackage{comment}
\usepackage{graphicx}

\usepackage{url}

\aclfinalcopy % Uncomment this line for the final submission
\title{Why we have switched from building full-fledged taxonomies \\ to \textit{simply} detecting hypernymy relations}

\author{Jose Camacho-Collados \\
  Department of Computer Science \\
   Sapienza University of Rome \\
  {\tt collados@di.uniroma1.it} \\
  }

\date{}

\begin{document}
\maketitle
\begin{abstract}
  The study of taxonomies and hypernymy relations has been extensive on the Natural Language Processing (NLP) literature. However, the evaluation of taxonomy learning approaches has been traditionally troublesome, as it mainly relies on \textit{ad-hoc} experiments which are hardly reproducible and manually expensive. Partly because of this, current research has been lately focusing on the hypernymy detection task. In this paper we reflect on this trend, analyzing issues related to current evaluation procedures. Finally, we propose two potential avenues for future work so that is-a relations and resources based on them play a more important role in downstream NLP applications.
\end{abstract}

\section{Introduction}
Taxonomies are hierarchical organizations of concepts from domains of knowledge. They generally constitute the backbone of ontologies \cite{Velardietal2013} and contribute to applications such as information search, retrieval, website navigation and records management \cite{Bordeaetal2015}, to name a few. In order to construct a taxonomy, a prior step of extracting hypernymy (is-a) relations between pairs of concepts needs to be performed. The result of this prior step is a list of edges which are later integrated into the taxonomic data structure. %(which, in addition, is unclear if should be only tree-like, or directed acyclic graphs would be appropriate as well).
This prior step has been shown to directly help in downstream applications such as Question Answering \cite{prager2008question,yahya2013robust} or semantic search \cite{hoffart2014stics}. Interestingly, the research attention seems to have shifted from building full-fledged taxonomies entirely or partly from scratch to this prior step of inducing hypernymy relations. We argue that this is largely due not only to the reduced complexity in collecting training data, but also to a more straightforward evaluation. There are no standard benchmarks for taxonomy evaluation and they mainly rely on non-replicable manual evaluation \cite{gupta-EtAl:2016:COLING2}. On the contrary, the hypernymy detection task is easier to evaluate as it counts with standard evaluation benchmarks which make the comparison of systems rather straightforward. However, previous work strongly questions the fitness of using a manually-crafted taxonomy like WordNet \cite{Miller1995} for evaluating systems that harvest terms and their corresponding hypernyms from text \cite{Hovyetal2009}. Additionally, there are other issues and questions related to the utility of hypernymy detection techniques in downstream applications that may arise due to current evaluation practices.

This discussion paper is structured as follows: we first introduce basic notions on the automatic acquisition of hierarchically-structured knowledge (Section \ref{background}). Then, we discuss current problems for evaluating taxonomy learning techniques, and explain the shift of current research towards the development of hypernymy detection models (Section \ref{issues}). Finally, we propose two lines of research for future work (Section \ref{conclusion}).

\section{Background}
 \label{background}

%Let us first introduce a bit the topic for those who are not very familiarized with it. Taxonomies are…. Hypernyms….

\begin{figure}[!h]
\begin{center}
\includegraphics[width=7.8cm]{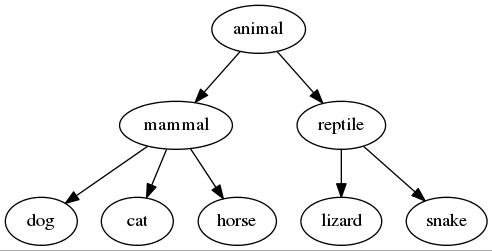}
\caption{A very simplified branch of an \textit{animal} taxonomy.}
\label{figureanimal}
\label{transport}
\end{center}
\end{figure}

%{\color{red} Add Nice figure of a taxonomy}
Representing domain-specific knowledge in the form of hierarchical concepts, i.e. taxonomies (see Figure \ref{figureanimal} for an example), has been a long-standing research problem.  %There are countless reasons of why it is desirable to develop systems that model a domain of knowledge automatically in terms of a semantic hierarchy. 
However, building resources like WordNet (or any domain-specific ontology) requires an almost prohibitive manual effort \cite{FountainandLapata2012}. %Reasons related to cost, consistency and coverage are among the most cited problems \cite{FountainandLapata2012}. 
Therefore, building taxonomies automatically (or semi-automatically) has attracted the interest of the NLP community. The process of automatically building taxonomies is usually divided into two main steps: finding hypernyms for concepts, which may constitute a field of research in itself (Section \ref{back:hyper}) and refining the structure into a taxonomy (Section \ref{back:taxo}).

\subsection{Hypernymy identification}
\label{back:hyper}

Hypernymy identification is generally split in two broad categories: pattern-based\footnote{\textit{Pattern-based} has also been referred to as \textit{path-based} \cite{Schwartzetal2016} or \textit{rule-based} \cite{NavigliandVelardi2010} in the literature.} and distributional. Pattern-based techniques exploit the co-ocurrences of the terms which compose the hypernymy relation in text corpora. Pattern-based methods have traditionally been based on the so-called \textit{Hearst's patterns} \cite{Hearst1992}, which are a set of lexico-syntactic patterns to identify hypernymy relations. Other approaches have built up on these patterns with the aid of various syntactic and statistical techniques \cite{Etzionietal2005,Snowetal2006,KozarevaandHovy2010}. Syntactic clues have also played an important role on more recent approaches exploiting supervised techniques \cite{NavigliandVelardi2010,BoellaandDiCaro2013}. 

The second branch for hypernymy identification exploits distributional models. These methods generally address the problem using supervised techniques over the distributional representations of the terms included in the hypernymy relation \cite{Baronietal2012,Rolleretal2014,weeds2014learning,Levyetal2015,Yuetal2015,roller-erk:2016:EMNLP2016}. Unlike pattern-based techniques, distributional models do not assume that terms and hypernyms co-occur in the same sentence. Recent approaches have also leveraged supervised distributional models exploiting a domain-aware transformation matrix between the vector spaces of terms and hypernyms \cite{Fuetal2014,EspinosaEMNLP2016,babeldomains:2017}. Finally, \newcite{Schwartzetal2016} proposed a LSTM-based architecture encoding both syntactical and distributional information which proved effective in the hypernymy detection task.

\subsection{Taxonomy learning}
\label{back:taxo}

As explained earlier, taxonomy learning is generally divided into two main phases: (1) identifying hypernymy relations from textual data, and (2) inducting a full-fledge taxonomy based on the relations extracted from the first step. In the literature, most taxonomies have been constructed by exploiting pattern-based approaches \cite{YangandCallan2009,KozarevaandHovy2010,Naviglietal2011,nakashole2012,Tuanetal2014,AlfaroneandDavis2015,Espinosa-Ankeetal2016AAAI,tiziano:2016aij,faralli2016linked,gupta-EtAl:2016:COLING2}. The second phase aims at forming the graph structure of a taxonomy. The techniques performed to achieve this goal vary from one model to another, but generally include the following steps: domain filtering, graph-based induction, and edge pruning and recovery \cite{Velardietal2013}.

%{\color{red} Add works on hypernyms }

%{\color{red} Add Nice figure of hypernyms}

\section{Analysis of Current Trends}
 \label{issues}

In this section we discuss current issues on the evaluation of taxonomies (Section \ref{tax-eval}) and the recent focus on the hypernymy detection task (Section \ref{hyper-det}).

\subsection{Taxonomy learning evaluation}
\label{tax-eval}

Traditional procedures to evaluate taxonomies have focused on measuring the quality of the edges, i.e., assessing the quality of the is-a relations \cite{ponzetto2011taxonomy,Flatietal2014}. This process typically consists of extracting a random sample of edges and manually labeling them by human judges. In addition to the manual effort required to perform this evaluation, this procedure is not easily replicable from taxonomy to taxonomy (which would most likely include different sets of concepts), and do not reflect the overall quality of a taxonomy \cite{gupta-EtAl:2016:COLING2}.

Moreover, some taxonomy learning approaches link their concepts to existing resources such as Wikipedia \cite{nakashole2012,Flatietal2014,tiziano:2016aij,gupta-EtAl:2016:COLING2}, BabelNet \cite{Espinosa-Ankeetal2016AAAI} or WordNet \cite{Suchanek2007,yamada2011extending,jurgens2015reserating}, while others remain at the word level \cite{KozarevaandHovy2010,AlfaroneandDavis2015}. This poses additional problems for evaluating the quality across different taxonomies\footnote{This issue is directly extensible to the hypernymy detection and hypernym discovery tasks as well (see Section \ref{hyper-det})}.

\subsection{Hypernymy detection}
\label{hyper-det}

Recently the research focus seems to have switched to the study of hypernymy relations only, which may be viewed as the first phase of taxonomy learning techniques (see Section \ref{back:taxo}) or a research field in itself. In particular, current approaches have been specializing on the hypernymy detection tasks \cite{santus2014chasing,weeds2014learning,Rolleretal2014,Schwartzetal2016,shwartz2017hypernymy}. The hypernymy detection task is a binary task consisting of, given a pair of words, deciding whether a hypernymic relation holds between them or not. In our view, this shift has occurred due to two main factors: 

\begin{enumerate}

    \item  The evaluation is definitely easier and more reliable since, as mentioned in Section \ref{tax-eval}, taxonomies generally rely on human-based evaluations that are hard to replicate. 
    
    \item  The rise of supervised distributional models and neural networks, which can be effectively used to detect hypernymy relations framing the task as a binary classification problem\footnote{This simplification of traditional tasks may be extensible to other areas of NLP, where complex tasks have been reduced to simpler classification problems. While this issue is clearly relevant and prone to be discussed, analyzing these trends through a more general perspective is out of the scope of this paper.}.

\end{enumerate}

The first reason is definitely a valid concern, as taxonomies have been proved difficult to evaluate, usually relying on ad-hoc manual evaluation which is hardly reproducible from one work to another (see Section \ref{tax-eval}). In fact, the creation of reliable hypernymy detection datasets have contributed to a rise of brand-new algorithms on the area. A popular dataset to evaluate hypernymy detection is BLESS \cite{Baronietal2011}, which includes additional relations such as meronymy or co-hyponymy. However, it is relatively small as it only contains 200 distinct target concepts. Other datasets directly rely on existing hand-crafted taxonomies like WordNet \cite{Snowetal2004,boleda2017eacl} or include additional resources such as Wikidata \cite{Vrandecicetal2014} or DBPedia \cite{auer2007dbpedia}, as in \newcite{Schwartzetal2016}.

However, what is the main utility of detecting a hypernymy relation between a pair of words? The most direct answer is that the main application is to be able to find hypernymy relations for a given concept, which is arguably the main practical feature in downstream applications, e.g. question answering (\textit{What is the longest river in Asia?}). However, if this is the main application, why do the evaluation focus on detecting hypernymy relations only? The step from \textit{detecting} hypernymy relations to \textit{discovering} or \textit{finding} hypernyms for a given concept is feasible but unfortunately not trivial. To the best of our knowledge there are no approaches which use a hypernymy detection system to “discover” hypernyms as a result. 

Additionally, considering that research on hypernyms is mainly envisaged to be used as a proxy for either constructing new taxonomies or integrating them into end-user applications, these automatic systems that are being evaluated on the hypernymy detection task using already existing taxonomies such as WordNet or Wikidata may not be reliable enough. One would definitely be more confident using the “gold-standard” WordNet or Wikidata on applications instead. Wikidata and especially WordNet clearly do not have a full coverage, especially on specialized domains. However, automatic systems are usually not evaluated outside these resources, which makes them unreliable (in the sense that they have not been tested) outside these resources. 

On a recent work, \newcite{EspinosaEMNLP2016}, presented a supervised domain-aware hypernym discovery system and evaluated it inside and outside Wikidata. Some results were encouraging but also showed that we are still far from having a reliable hypernym discovery system which could replace existing taxonomies in many domains. Further research should focus on constructing better benchmarks and developing methods which are ready to be deployed in downstream applications by extending or/and accurately replacing the is-a edges of current hand-crafted taxonomies, particularly on specialized domains.

\section{Conclusion and Future Work}
 \label{conclusion}
 
 In this paper we have discussed the current state of hypernym and taxonomy research. We have particularly focused on some issues arisen from standard evaluation practices present. Based on the main insights extracted from this discussion, we present two possible lines for future research: improvement of current evaluation practices for taxonomy evaluation (Section \ref{concl3}) and the development of systems for the hypernymy discovery task along with the creation of new challenging benchmarks (Section \ref{concl1}). %, and construction of new hypernymy discovery benchmarks (Section \ref{concl2}).

\subsection{Improvement of taxonomy evaluation procedures} 
\label{concl3}

As explained in Section \ref{tax-eval}, the evaluation of taxonomy learning techniques have been troublesome for a number of reasons, being one of them the importance given to measuring the accuracy of their is-a relations. Taxonomies should certainly evaluate their is-a relations quantitatively but the manual effort and non-replicability of these experiments make them undesirable. One possible solution to this problem could be to rely on the hypernym discovery task (see the following section). However, this is not enough to provide a global evaluation of the taxonomy. Further research should focus on developing reliable evaluation frameworks for the different features of a taxonomy, following the line of \newcite{gupta-EtAl:2016:COLING2}. \newcite{gupta-EtAl:2016:COLING2} who proposed a comprehensive evaluation framework going beyond the edge-level by additionally evaluating the granularity and generalization paths of a taxonomy. 

As far as taxonomy learning approaches that link their concepts to other existing taxonomies are concerned \cite{yamada2011extending,jurgens2015reserating}, the SemEval 2016 task on WordNet taxonomy enrichment \cite{jurgens2016semeval} may constitute a suitable evaluation framework.

%\subsection{Development of hypernym discovery systems} 
%\label{concl1}

\subsection{Construction of new challenging benchmarks for hypernym discovery} 
\label{concl1}

As discussed in Section \ref{hyper-det}, we argue that future research should pay a renewed attention to the hypernym discovery in addition to the hypernymy detection task. Hypernym discovery may be viewed as the first step of taxonomy learning techniques and constitutes a research field in itself. Systems may be tested on extensions of existing hypernymy detection datasets based on resources like Wikidata or WordNet, e.g. \cite{Snowetal2004,Schwartzetal2016}. Standard datasets are reduced to the hypernymy detection binary classification task, i.e. the system should decide whether \textit{dog-animal} or \textit{dog-mammal} constitute a hypernymy relation or not. Instead, using existing taxonomies and a similar procedure, a dataset could be created for the \textit{hypernym discovery} task, i.e., given the word \textit{dog}, the system should be able to discover its hypernyms \textit{mammal} or \textit{animal}, among others. For the evaluation we can \textit{borrow} some traditional information retrieval measures such as Mean Reciprocal Rank (MRR), Mean Average Precision (MAP), R-Precision (R-P) or Precicion at $k$ (P@$k$) (see \newcite{Bianetal2008} for more details on these measures). %We could argue that this task is definitely more challenging and more realistic when thinking of end-user applications in comparison with the binary hypernymy detection task. 

%\subsection{Construction of challenging hypernym discovery benchmarks}  
%\label{concl2}

In order to assess the reliability of hypernym detection systems, new benchmarks would be required. Apart from current hypernymy detection datasets based on existing lexical resources which can be re-framed for the hypernymy discovery task, we should construct challenging datasets independent from existing lexical resources, or constructed on top of these resources. This will show the potential of our systems to go beyond existing manually-crafted taxonomies. There are some existing domain-specific datasets from SemEval \cite{Bordeaetal2015,bordea2016semeval} which are rarely used in practice but may constitute an interesting starting point for this new evaluation branch. As an extrinsic evaluation it may be useful to construct datasets which are a direct proxy of downstream NLP applications such as question answering or information retrieval.

Additionally, it has shown by \newcite{boleda2017eacl} that concepts and entities/instances behave differently with respect to hypernymy relations. \textit{Laika-dog} (\textit{Laika} being a instance) and \textit{dog-mammal} (\textit{dog} being a concept) are, respectively, two examples of instantiation and hypernymy relation. These two different kinds of relation, which are often interchangeable in the literature, may be further introduced in these datasets.

\section*{Acknowledgments}

Jose Camacho-Collados is supported by a Google Doctoral Fellowship in Natural Language Processing. I would also like to thank Luis Espinosa-Anke for our interesting discussions on this topic.

%The acknowledgments should go immediately before the references.  Do
%not number the acknowledgments section. Do not include this section
%when submitting your paper for review.

%\bibliography{acl2017}
\bibliography{acl2017}
\bibliographystyle{acl_natbib}

\end{document}